\documentclass[runningheads]{llncs}
\usepackage{amsmath,amssymb,amsfonts}
\usepackage{algorithmic}

\usepackage{textcomp}
\usepackage{xcolor}

\usepackage[T1]{fontenc}
\usepackage[utf8]{inputenc}
\usepackage[english]{babel}
\usepackage[bookmarks=false]{hyperref}       
\usepackage{url}            
\usepackage{booktabs}       
\usepackage{amsfonts}       
\usepackage{nicefrac}       
\usepackage{microtype}      
\usepackage{multicol}
\usepackage{lipsum}
\usepackage{csquotes}
\usepackage{graphicx}
\usepackage{caption}
\usepackage{subcaption}
\usepackage{multirow}
\usepackage{amssymb}
\usepackage[export]{adjustbox}
\usepackage[misc,geometry]{ifsym}

\usepackage{pifont}
\newcommand{\xmark}{\ding{55}}

\newcommand\modelname{Faster DAN}

\newcommand{\mb}[1]{\boldsymbol{#1}}

\newif \ifanonymous
\anonymousfalse

\begin{document}

\title{\modelname{}: Multi-target Queries with Document Positional Encoding for End-to-end Handwritten Document Recognition}



\ifanonymous
\else
\author{Denis Coquenet\inst{1}(\Letter)\orcidID{0000-0001-5203-9423} \and
Clément Chatelain\inst{2,3}\orcidID{0000-0001-8377-0630} \and
Thierry Paquet\inst{2,4}\orcidID{0000-0002-2044-7542}}

\institute{Conservatoire National des Arts et Métiers, CEDRIC, Paris, France \and LITIS Laboratory - EA 4108, France \and Rouen University, France \and INSA of Rouen, France\\
    \email{denis.coquenet@lecnam.net}\\
    \email{\{clement.chatelain, thierry.paquet\}@litislab.eu}
}
\fi

\maketitle
\thispagestyle{plain}
\pagestyle{plain}

\begin{abstract}
Recent advances in handwritten text recognition enabled to recognize whole documents in an end-to-end way: the Document Attention Network (DAN) \cite{Coquenet2022b} recognizes the characters one after the other through an attention-based prediction process until reaching the end of the document. 
However, this autoregressive process leads to inference that cannot benefit from any parallelization optimization. In this paper, we propose \modelname{}, a two-step strategy to speed up the recognition process at prediction time: the model predicts the first character of each text line in the document, and then completes all the text lines in parallel through multi-target queries and a specific document positional encoding scheme. \modelname{} reaches competitive results compared to standard DAN, while being at least 4 times faster on whole single-page and double-page images of the RIMES 2009, READ 2016 and MAURDOR datasets. Source code and trained model weights are available at \url{https://github.com/FactoDeepLearning/FasterDAN}.
\keywords{Handwritten Document Recognition, Document Layout Analysis, Handwritten Text Recognition, Transformer}
\end{abstract}

\section{Introduction}

Unconstrained offline handwritten text recognition has been studied for decades now. Until recently, all the proposed approaches were focused on recognizing the text from cropped parts (text regions) of the original document, leading to a sequential multistep approach, namely text region segmentation, ordering and recognition.
Numerous advances enabled to extend the recognition stage to handle increasingly complex inputs. In the 90's, the use of Hidden Markov Model (HMM) enabled to go from isolated character recognition \cite{LeCun89} to multi-character (word or line) recognition \cite{Gilloux1995,Yacoubi1999}. Thereafter, the democratization of deep neural networks, combined with the Connectionist Temporal Classification (CTC) loss \cite{Graves2006}, made the line-level approach the standard framework to handle handwritten document recognition \cite{Graves2008,Voigtlaender2016,Puigcerver2017,Ptucha2019,Coquenet2019,Coquenet2020,Yousef2020b}.

\begin{figure}[h!]
    \centering
    \includegraphics[width=0.7\textwidth]{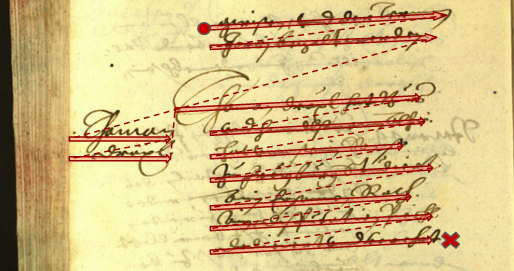}
    \includegraphics[width=0.7\textwidth]{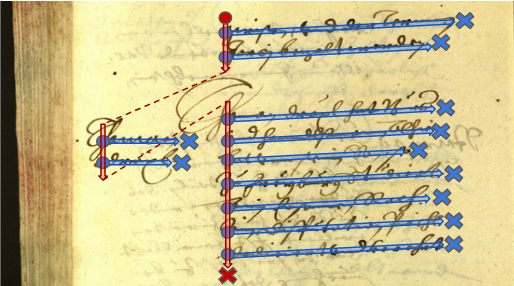}
    \caption{Reading order comparison between DAN (top) and \modelname{} (bottom). Circles and crosses represent the start and the end of a pass, respectively. The first pass is showed in red, and the second one in blue. The DAN (top) sequentially predicts the characters of the whole documents in a single pass. The \modelname{} first predicts the first character of each line (as well as the layout tokens), and then predicts the remaining of all the text lines in parallel, in a second pass.}
    \label{fig:reading-order}
\end{figure}

More recently, few advanced works focused on text recognition  at paragraph level \cite{Bluche2016,Bluche2017,Yousef2020a,Coquenet2021}, reaching similar performance compared to line-level recognition \cite{Coquenet2023}. However, whether it is at character, word, line or paragraph level, this three-step paradigm has many drawbacks: the errors accumulate from one step to another, additional physical segmentation annotations are required to train the segmentation step, the use of a rule-based ordering stage is limited for documents with a complex layout, and the stages are performed independently, so they cannot benefit from one another.

\ifanonymous
\else 
Based on these observations, we proposed in \cite{Coquenet2022b} a new end-to-end paradigm named Handwritten Document Recognition (HDR).
\fi
It aims at serializing documents in an XML-way, combining both Handwritten Text Recognition (HTR) and Document Layout Analysis (DLA), through layout XML-markups. 
\ifanonymous
\else
We proposed the Document Attention Network (DAN) \cite{Coquenet2022b} to tackle HDR.
\fi
It is made up of a Fully Convolutional Network (FCN) encoder to extract features from the input image, and a transformer \cite{Vaswani2017} decoder to recurrently predict the different character and layout tokens. The DAN reached competitive results, recognizing both text and layout at page or double-page levels, compared to state-of-the-art line-level or paragraph-level approaches. The main drawback of the DAN is about its autoregressive character-level prediction process, which leads to high prediction times (a few seconds per document image). 

In this paper, we propose \modelname{}, a novel approach to significantly reduce the prediction time of end-to-end HDR, without impacting the training time. This approach is based on a new document positional encoding whose aim is to inject the line membership information to each predicted character. In this way, we can parallelize the recognition of the text lines still using a single model, while reducing the total number of iterations. The \modelname{} relies on a two-step prediction process: a first step is dedicated to the prediction of the layout tokens, as well as the first character of each text line; all the text lines are then recognized in parallel in the second stage through multi-target transformer queries. This is illustrated in Figure \ref{fig:reading-order}.

We show that the \modelname{} reaches competitive results compared to the original DAN, while being at least 4 times faster on three public datasets: READ 2016, RIMES 2009 and MAURDOR.

This paper is organized as follows. Section \ref{related_work} is dedicated to the related works. DAN background is presented in Section \ref{background}. We detail the proposed approach in Section \ref{faster-dan}. Section \ref{experiments} presents the experimental environment and the results. We draw the conclusion in Section \ref{conclusion}.
\section{Related Works}
\label{related_work}

Nowadays, the most popular HTR framework is made up of three stages: the input document image is segmented into text line crops, which are then ordered and recognized. Indeed, the concept of text line is widely used as a building block in many works, and has been studied from different angles. 

The text line has mostly been studied from the physical point of view: the majority of the works focused on predicting text line bounding boxes, either through a pixel-by-pixel classification task \cite{Renton2018,Oliveira2018,Boillet2022} or through an object-detection approach \cite{Chung2019,Carbonell2020}.
Detecting the start-of-line information was also studied as part of the segmentation stage. In \cite{Moysset2017}, a model is trained to predict the coordinates of the bottom-left corner of each text line, as well as their height. Similarly, in \cite{Wigington2018,Wigington2019}, the authors considered the prediction of the start-of-line coordinates as an object detection task, using a region proposal network. Scale and rotation values are also associated to each text line to handle monotonic slanted lines.
Contrary to these works, the \modelname{} strategy we propose only relies on language supervision: we do not need any additional physical annotations.

Recent works proposed to perform the recognition step at paragraph level \cite{Bluche2016,Bluche2017,Coquenet2021,Yousef2020b}. Although not relying on raw physical text line annotations, most paragraph-level text recognition works take advantage of the physical properties of text lines in single-column layout: the whole horizontal axis is associated to a text line, no matter its length. The authors of \cite{Yousef2020a} and \cite{Coquenet2021} concatenate the representation of the different text lines, or the text line predictions, respectively, to get back to a one-dimensional alignment problem. In \cite{Bluche2016,Coquenet2023}, the text lines are processed recurrently through a line-level attention mechanism. 

Another approach to deal with multi-line images consists in relying on an autoregressive character-level prediction process \cite{Bluche2017,Singh2021,Rouhou2021,Coquenet2022b}. This time, the notion of text line is limited to the use of a dedicated line break token, used as any other character token. This way, this approach is no longer limited to single-column document. This strategy is also used in \cite{Kim2022} for visual question-answering, information extraction or classification of documents, the OCR task being reduced to pretraining.
\ifanonymous
\else
In \cite{Coquenet2022b}, we proposed the Document Attention Network
\fi
to tackle Handwritten Document Recognition, by predicting opening and closing layout markup tokens in an XML way: all the character and layout tokens are sequentially and indifferently predicted, leading to hundreds or even thousands of iterations for single-page or double-page document images. It results in long prediction times: approximately one second for 100 characters on a single GPU V100.

In this paper, we propose to speed up the prediction of this latter approach by reading text lines in parallel. This way, we take the best of both worlds: we can deal with documents with complex layout through this character-level attention, and we use the concept of text line more directly through the prediction of the first character of each line and by using a dedicated document positional encoding scheme, but without using any additional physical annotations.

\section{DAN background}
\label{background}
\ifanonymous
\else
We proposed the Document Attention Network (DAN) in \cite{Coquenet2022b} for the task of Handwritten Document Recognition.
\fi
It takes an input image of a whole document $\mb{X} \in \mathbb{R}^{H_\text{i} \times W_\text{i} \times C_\text{i}}$, where $H_\text{i}$, $W_\text{i}$, $C_\text{i}$ are the height, the width and the number of channels, respectively. It outputs the associated XML-like serialized representation $\hat{\mb{y}}$, \textit{i.e.}, a sequence of tokens, each token $\hat{\mb{y}}_i$ representing either a layout markup or a character among an alphabet $\mathcal{A}^*$. For an input document represented by $N$ tokens, we can note the expected output sequence as $\mb{y}^* \in {\mathcal{A}^*}^N$.
For instance, a three-line document, split into two paragraphs, could be represented as:

{\centering 
<D><P>Line 1\textbackslash nLine 2</P><P>Line 3</P></D> \\} 
where <D> and <P> corresponds to document and paragraph markups, respectively.

The DAN is made up of two main components. An FCN encoder is used to extract 2D features $\mb{f}^\text{2D} \in \mathbb{R}^{H \times W \times d}$ from the input image $\mb{X}$, with $H=\frac{H_\text{i}}{32}$, $W=\frac{W_\text{i}}{8}$ and $d=256$.
A Transformer decoder is used to iteratively predict the tokens $\hat{\mb{y}}_i$. To this aim, a special start-of-transcription token is used to initialize the prediction ($\hat{\mb{y}}_0=\text{<sot>}$) and a special end-of-transcription token is added to the ground truth to stop it.
This way, the new target sequence is $\mb{y} \in \mathcal{A}^{N+1}$ with $\mb{y}_{N+1}=\text{<eot>}$ and $\mathcal{A} = \mathcal{A}^* \cup \{ \text{<eot>} \}$. During inference, a maximum number of iterations $N_\text{max}=3,000$ is fixed in case of the <eot> token is not predicted.

The transformer attention mechanism being invariant to the order of its input sequences, positional encoding is added to inject the positional information: 2D positional encoding $\mb{P}^\text{2D} \in \mathbb{R}^{H \times W \times d}$ for the 2D features of the image, and 1D positional encoding $\mb{P}^\text{1D} \in \mathbb{R}^{N_\text{max} \times d}$ for the previously predicted tokens. Both positional encodings are defined as a fixed encoding based on sine and cosine functions with different frequencies, as proposed in the original Transformer paper \cite{Vaswani2017}. The image features are flattened afterward, for transformer needs:
\begin{equation}
    \mb{f}^\text{1D} = \text{flatten}(\mb{f}^\text{2D} + \mb{P}^\text{2D}).
\end{equation}

The DAN can be seen under the prism of the question-answering paradigm. At iteration $t$, the question corresponds to the previously predicted tokens $\mb{\hat{y}}^{\mb{t}}=[\hat{\mb{y}}_0, ..., \hat{\mb{y}}_{t-1}]$, referred to as \textit{context} in this work, and the answer is the next token $\hat{\mb{y}}_{t}$. Formally, the tokens are first embedded through a learnable matrix $\mb{E} \in \mathbb{R}^{(|\mathcal{A}|+1) \times d}$ (+1 for the <sot> token), leading to $\mb{e}^{\mb{t}} = [\mb{e}_0, ..., \mb{e}_{t-1}]$, with $\mb{e}_i = \mb{E}_{\hat{\mb{y}}_i}$ ($\in \mathbb{R}^{d}$). Positional embedding is then added to get the transformer input query $\mb{q}^{\mb{t}} = [\mb{q}_0, ..., \mb{q}_{t-1}]$ with $\mb{q}_i = \mb{e}_i+\mb{P}^\text{1D}_i$.

The transformer's self-attention and cross-attention mechanisms compute an output $\mb{o}_i \in \mathbb{R}^{d}$ for each query input $\mb{q}_i$ by comparing them with the other query tokens, and with the image features $\mb{f}^\text{1D}$, respectively. 
Formally, 
\begin{equation}
\mb{o}^{\mb{t}} = [\mb{o}_0, ..., \mb{o}_{t-1}] = \text{decoder}(\mb{q}^{\mb{t}}, \mb{f}^\text{1D}),
\label{eq:decoder}
\end{equation}
where the decoder corresponds to a stack of 8 standard transformer decoder layers \cite{Vaswani2017}. This process being autoregressive, the query at position $i$ can only attend to positions from $0$ to $i$. In addition, the intermediate computations are preserved for each layer from one iteration to another in order to avoid computing the same output multiple times. 

A score $\mb{s}^t_i$  is computed for each token $i$ of the alphabet $\mathcal{A}$ using a single densely-connected layer of weights $\mb{W}_p$ ($\mb{s}^t \in \mathbb{R}^{|\mathcal{A}|}$):
\begin{equation}
\mb{s}^t = \mb{W}_p \cdot \mb{o}_{t-1}.
\end{equation}
Probabilities are obtained through softmax activation: $\mb{p}^t_i = \frac{\exp{\mb{s}^t_i}}{\sum_j \exp{\mb{s}^t_j}}$.
The predicted token at iteration $t$ is the one whose probability is maximum: 
\begin{equation}
    \hat{\mb{y}}_t=\text{arg}\max(\mb{p}^t).
    \label{eq:argmax}
\end{equation}

The model is trained in an end-to-end fashion using the cross-entropy loss over the sequence of tokens: 
\begin{equation}
    \mathcal{L}_\text{DAN} = \sum_{t=1}^{N+1} \mathcal{L}_\text{CE}(\mb{y}_t, \mb{p}^t).
\end{equation}

This autoregressive process can be parallelized during training through teacher forcing, but this is not possible during inference. That is why we propose the \modelname{} strategy.

\section{\modelname{}}
\label{faster-dan}

The standard character-level attention-based approach for HTR is to sequentially recognize all the characters $\mb{y}_i$ of the whole input image $\mb{X}$. 
This way the number of iterations, thus the prediction time, grows linearly with the number of characters in the document. This may be negligible for isolated text line images, for which the image feature extraction stage is predominant, but this becomes significant for whole page images (around one second for 100 characters on a GPU V100).

We propose the \modelname{}, a novel approach for Handwritten Document Recognition, to noticeably reduce the prediction time. The goal is to take advantage of the line-based structure of documents to parallelize the recognition of the text lines. Considering the layout markups and the <eot> tokens as lines by themselves (of unit length), we can rewrite the target sequence as $\mb{y}=\text{concatenate}(\mb{y}^1, ..., \mb{y}^L)$ where $L$ is the number of lines in the document and $\mb{y}^j \in \mathcal{A}^{n_j}$ represent the different text lines ($\mb{y}^{j}_i$ is the character $i$ of  line $j$). 

Using one model per line is prohibitive in terms of GPU memory consumption. Instead, the parallelization is carried out among a single model which processes multi-target queries through masking in the second pass. This is feasible thanks to the dedicated document positional encoding scheme we propose. 
It is important to note that the proposed approach is not specific to the DAN architecture. It could be used with any attention-based HDR model. However, to our knowledge, the only available end-to-end HDR model is the DAN.

\subsubsection*{Reading lines in parallel}

Parallelizing the recognition faces two main challenges: detecting all the text lines, and recognizing them in parallel through transformer queries without mixing them. Moreover, since our goal is to perform HDR, and not only HTR, we also need to recognize the layout entities.

\begin{figure}[h!]
    \centering
    \begin{subfigure}{\textwidth}
    \includegraphics[width=\textwidth]{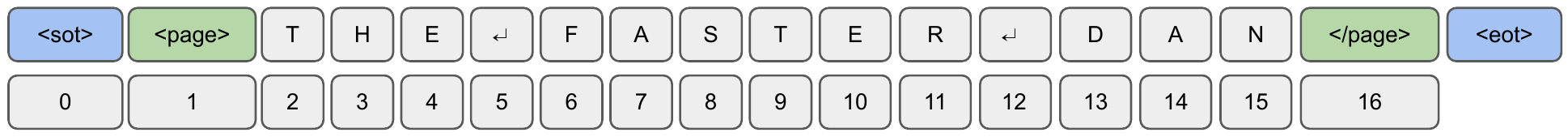}
    \caption{DAN single-pass prediction process}
    \end{subfigure}
    
    \begin{subfigure}{\textwidth}
    \includegraphics[width=\textwidth]{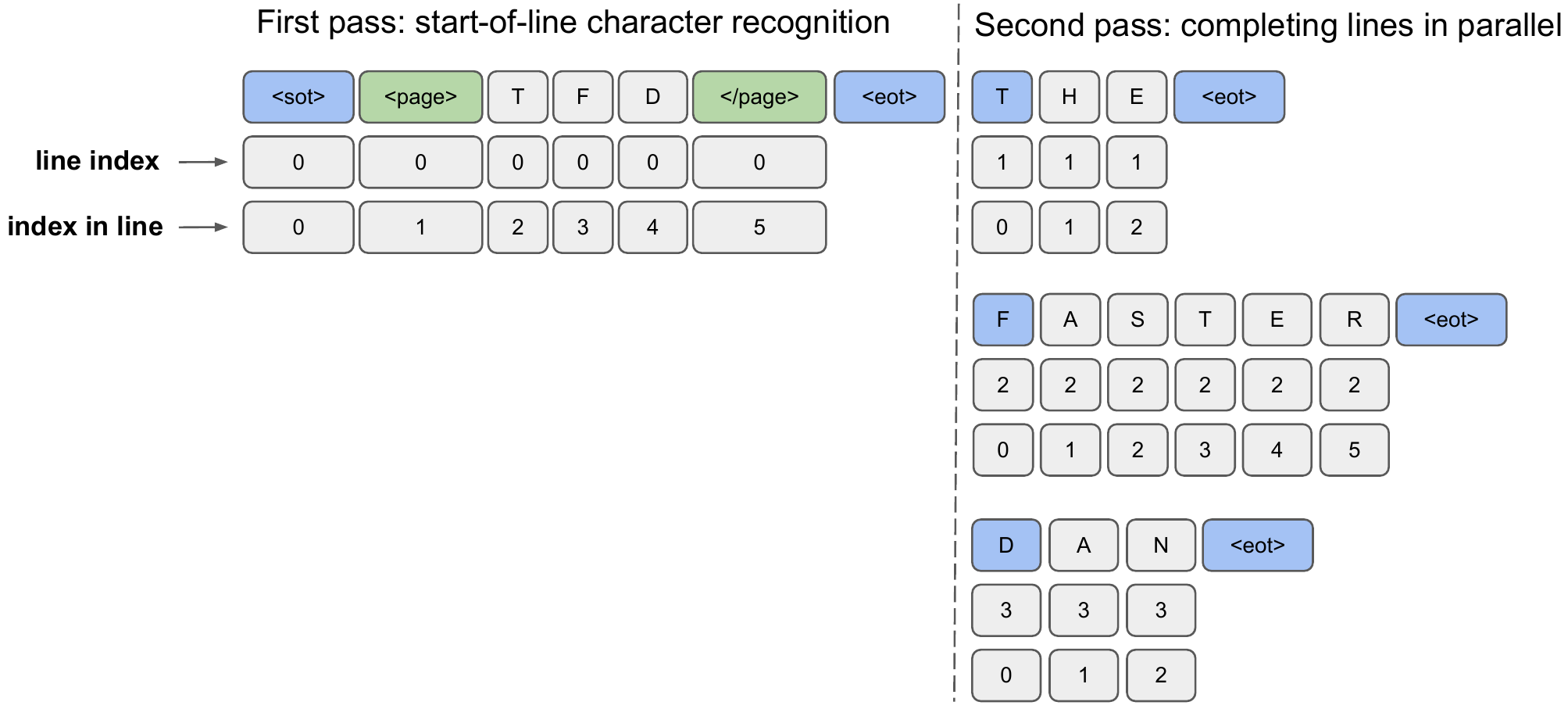}
    \caption{\modelname{} two-pass prediction process}
    \label{fig:gt-fdan}
    \end{subfigure}
    
    \caption{Comparison of the prediction process and positional encoding scheme between DAN and \modelname{}. This illustrates the example of a document input with three one-word text lines. The DAN associates a unique positional value for each token, which continues from one text line to the next. The \modelname{} uses two positional values: the index of the text line and the position of the token in this text line. Special (start and end) tokens are in blue and layouts tokens are in green.}
    \label{fig:gt}
\end{figure}

To tackle these issues, we opted for a two-pass process, as illustrated in Figure \ref{fig:gt-fdan}. In a first pass, the model sequentially predicts the layout tokens as well as the first character of each text lines, solving both layout recognition and text line detection. Then, the different text lines are completed in parallel based on their previously predicted first character. To this end, it is crucial to determine which token belongs to which line.

\subsubsection*{Document positional encoding}

To parallelize the recognition of the text lines, we propose a new positional encoding scheme, as shown in Figure \ref{fig:gt}. We associate to each predicted token $\hat{\mb{y}}^j_i$ (with $\hat{\mb{y}}^0_0 = \text{<sot>}$) two 1D positional embedding: one for the index of the line, and the other one for the index of the token in the line, leading to the global positional embedding $\mb{P}^\text{doc} \in \mathbb{R}^{l_\text{max} \times n_\text{max} \times d}$, where $l_\text{max}$ is the maximum number of line and $n_\text{max}$ is the maximum number of characters per line. $\mb{y}^{j}_i$ is associated to: 
\begin{equation}
    \mb{P}^\text{doc}_{j,i} = \text{concatenate}(\mb{P}^\text{1D'}_{j}, \mb{P}^\text{1D'}_{i}),
\end{equation}
where $\mb{P}^\text{1D'}$ is equivalent to $\mb{P}^\text{1D}$ but encoded on half channels ($\mb{P}^\text{1D'}_i \in \mathbb{R}^{d/2}$). The transformer input queries become $\mb{q}^j_i = \mb{E}_{\hat{y}^j_i} + \mb{P}^\text{doc}_{j,i}$.
The idea of injecting the line information was already used in \cite{Singh2021}, but it was computed as a ratio with an arbitrary maximum number of lines, and concatenated to the token embedding directly. In addition, the position of the tokens was absolute, and not relative to the current text line, as for the standard DAN. 

\subsubsection{First pass}
The \modelname{} follows the standard autoregressive process to predict the first token $\hat{\mb{y}}^{j}_0$ of each line $j$ based on Equations \ref{eq:decoder} to \ref{eq:argmax}. At iteration $t$,  $\mb{q}^{\mb{t}}= [\mb{q}^0_0, ..., \mb{q}^{t-1}_0]$.

\subsubsection{Second pass}
The standard Transformer decoding process is to give a sequence of query tokens $\mb{q}^{\mb{t}}$ as input and keep the output corresponding to the last token only ($\mb{o}_{t-1}$), as single output for iteration $t$. Instead, the output of the last token of each line $\mb{o}^{j}_{t-1}$ are kept in this second pass. We refer to this as multi-target queries.
$\hat{\mb{y}}^{j}_0$ are duplicated into $\hat{\mb{y}}^{j}_1$ to initiate the second pass; the modification of the associated position in line (from 0 to 1) indicates to the model a change of expected behavior: from the prediction of the first token of the next line to the prediction of the next token of the current line.
By setting $\mb{q}^{\mb{t}} = [ \mb{q}^0_0, ...,  \mb{q}^0_{t-1}, ..., \mb{q}^L_0, ..., \mb{q}^L_{t-1}]$ (the $t$ first tokens of all the lines), we obtain $\mb{o}^{\mb{t}} = [ \mb{o}^0_0, ...,  \mb{o}^0_{t-1}, ..., \mb{o}^L_0, ..., \mb{o}^L_{t-1}]$ through Equation \ref{eq:decoder}.
In this way, the $t^\text{th}$ tokens of each line $j$ are computed in a single iteration $t$:
\begin{equation}
    \hat{\mb{y}}^j_t = \text{arg}\max(\mb{W}_p \cdot \mb{o}^{j}_{t-1}).
\end{equation}
Extra tokens ($\hat{\mb{y}}^j_i \text{ with } i>n_j$) are discarded through masking.

\subsubsection*{Context exploitation}
\begin{figure}[h!]
    \centering
    \begin{subfigure}{\textwidth}
    \includegraphics[width=\textwidth]{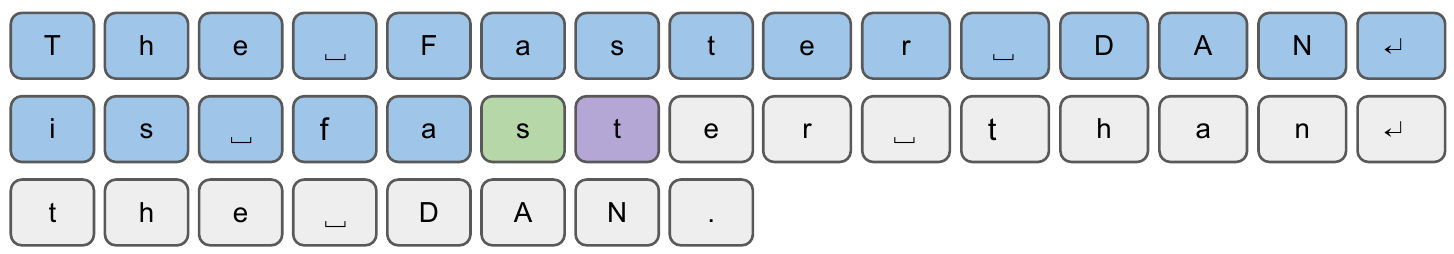}
    \caption{Context used by the DAN}
    \end{subfigure}
    
    \begin{subfigure}{\textwidth}
    \includegraphics[width=\textwidth]{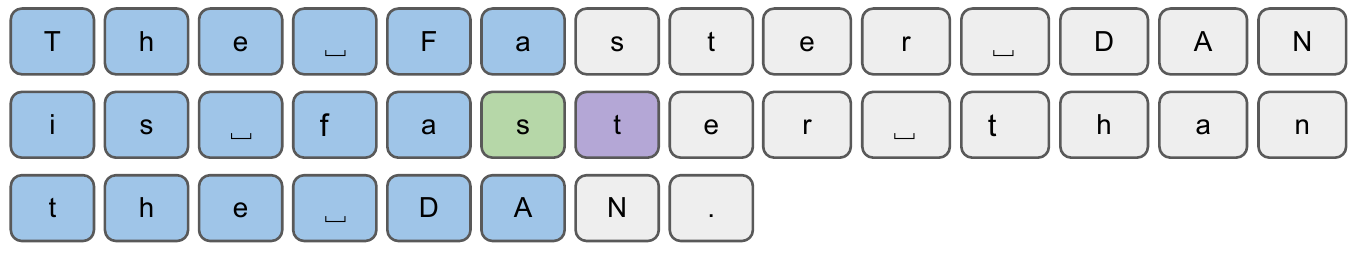}
    \caption{Context used by the \modelname{}}
    \end{subfigure}
    
    \caption{Context comparison between DAN and \modelname{}. The colored cells represent the current character to predict (in purple), the previously predicted tokens \textit{i.e.} the context (in blue and green), the token used for the prediction (in green), and the remaining characters to recognize (in gray).}
    \label{fig:context}
\end{figure}

The naive approach to recognize the text lines in parallel would be to recognize them independently, by applying a mask to discard the tokens from all the other text lines. It means that $\mb{q}^j_i$ could only attend to line $j$ (itself) and position $0$ to $i$ in that line. However, this would lead to an important loss of context. 
Instead, we propose to take advantage of all the partially predicted text lines: $\mb{q}^j_i$ can attend to all lines, from $0$ to $L$, and from position $0$ to $i$ in those lines, this is illustrated in Figure \ref{fig:context}

The major drawback of parallelizing the line recognition, compared to purely sequential recognition, is the loss of context. Indeed, the standard DAN benefits from all the past context during prediction: this is partially available for the \modelname{} since the past context is limited to the beginning of the text lines. In this way, it becomes harder for the model to focus on the correct text part, especially for very short contexts. Indeed, a sequence of characters may appear several times in a document, especially if this sequence is short, \textit{e.g.,} at the beginning of the recognition process. 
We counterbalance the loss of context from past by combining partial context from both past and future. We show the impact of this approach in Section \ref{section:ablation}.

\subsubsection*{Training and inference}
The model is trained over the target sequence using the cross-entropy loss:

\begin{equation}
\mathcal{L} = \sum_{j=1}^{L} \sum_{\substack{i =0 \\ i \neq 1}}^{n_j} \mathcal{L}_\text{CE}(\mb{y}^j_i, \mb{p}^j_i).
\end{equation}
It has to be noted that the training time is not impacted by this two-step decoding strategy since the whole expected sequence prediction (from both passes) is trained in parallel through teacher forcing, with appropriate masks.

During inference, the \modelname{} reduces the number of iterations $I$ from
\begin{equation*}
    I_\text{DAN} = \displaystyle \sum_{j=1}^{L} n_j = N+1
    \;\;\; \text{ to } \;\;\;
    I_\text{FasterDAN} = L + \max_j(n_j),
\end{equation*}
by considering the line breaks as belonging to the lines.
For example, 25 text lines of 50 characters, structured according to 3 layout entities, leads to 1,251 iterations for the DAN, and only 76 iterations for the proposed \modelname{}.

\section{Experimental study}
\label{experiments}


\subsection{Datasets}
\begin{figure}[h!]
    \captionsetup[subfigure]{justification=centering}
    \centering
    \begin{subfigure}{0.3\textwidth}
        \centering
        \includegraphics[width=\textwidth,frame]{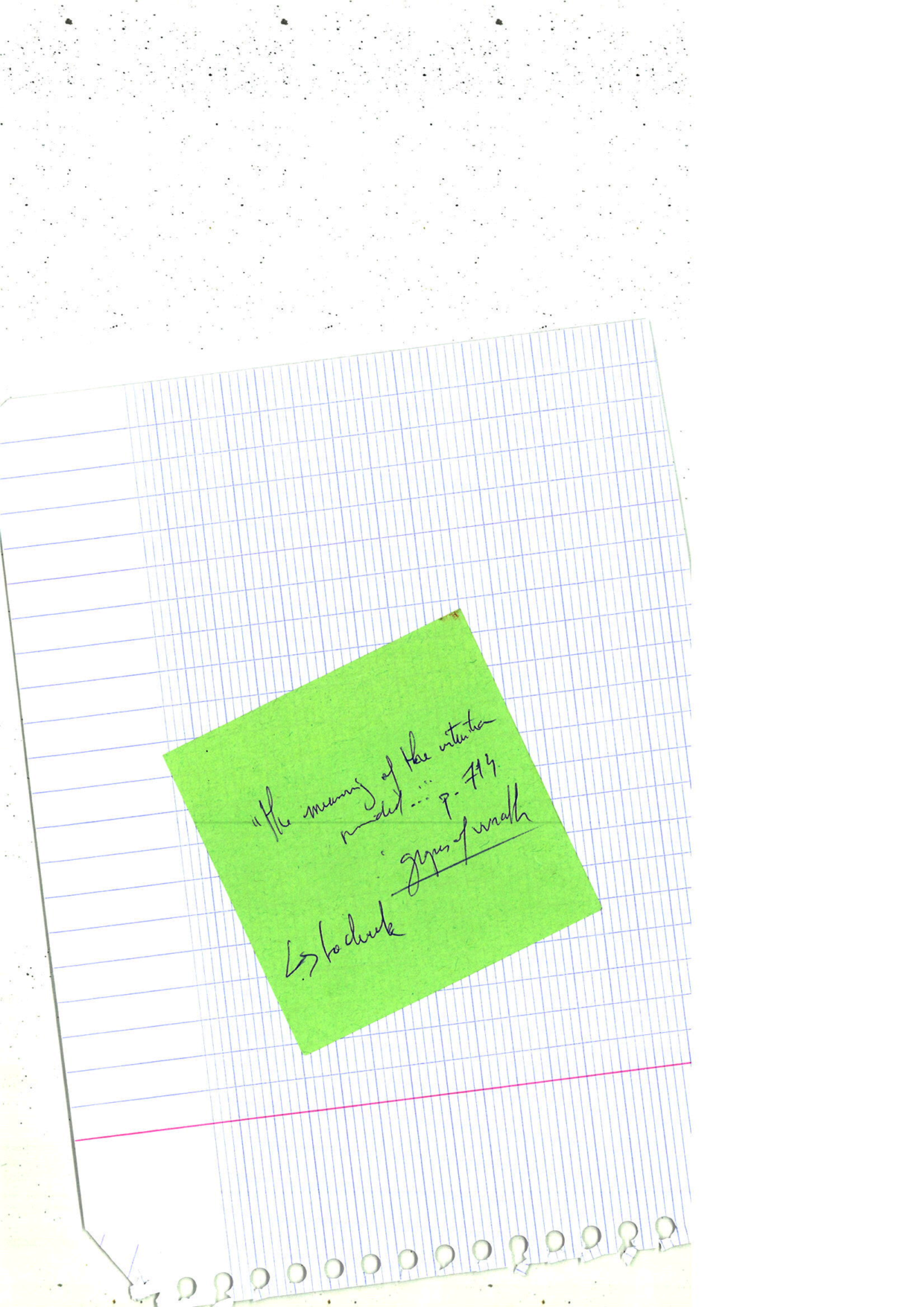}
        \caption{MAURDOR C3} 
    \end{subfigure}
    ~
    \begin{subfigure}{0.3\textwidth}
        \centering
        \includegraphics[width=\textwidth,frame]{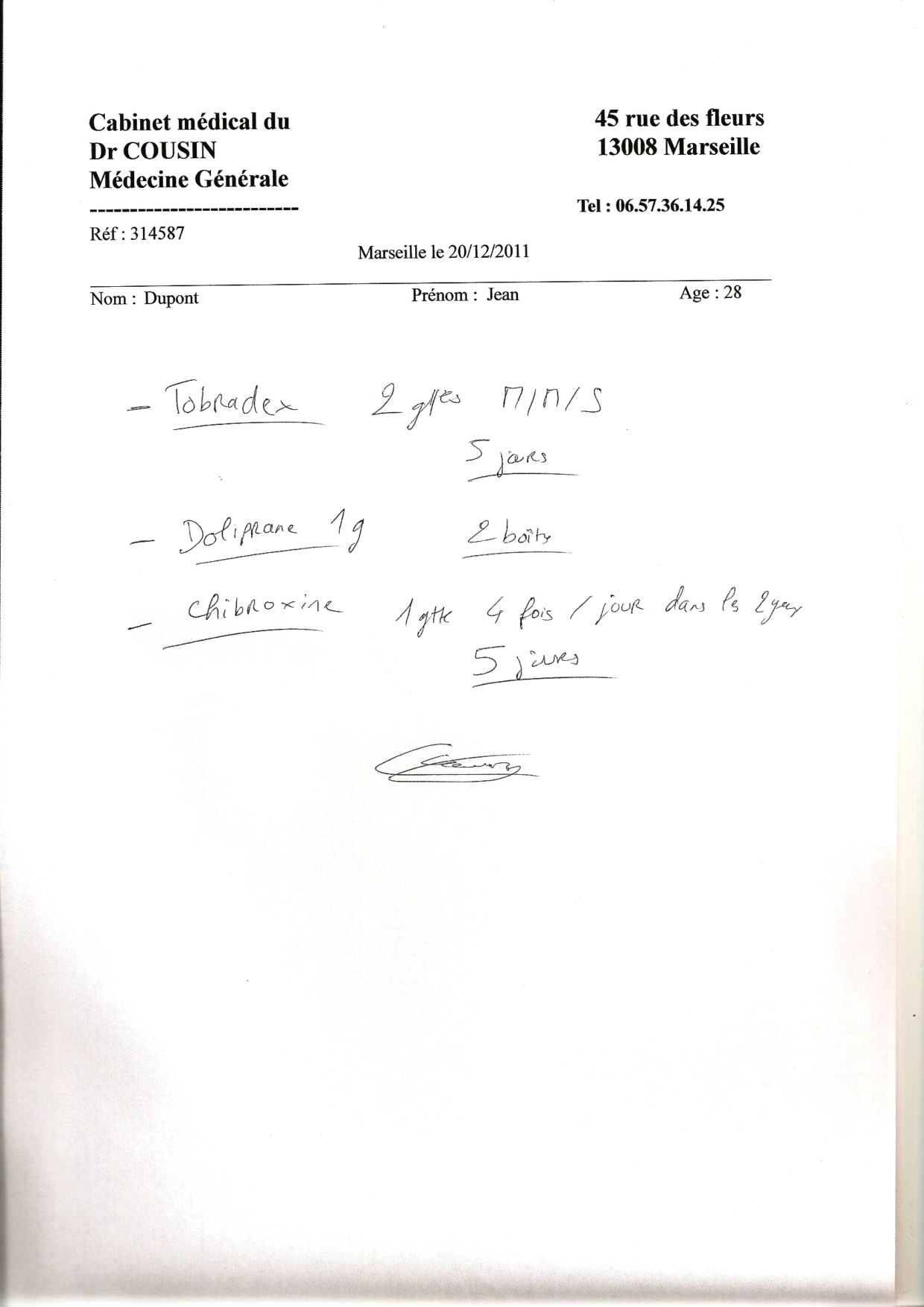}
        \caption{MAURDOR C4} 
    \end{subfigure}
    ~
    \begin{subfigure}{0.3\textwidth}
        \centering
        \includegraphics[width=\textwidth,frame]{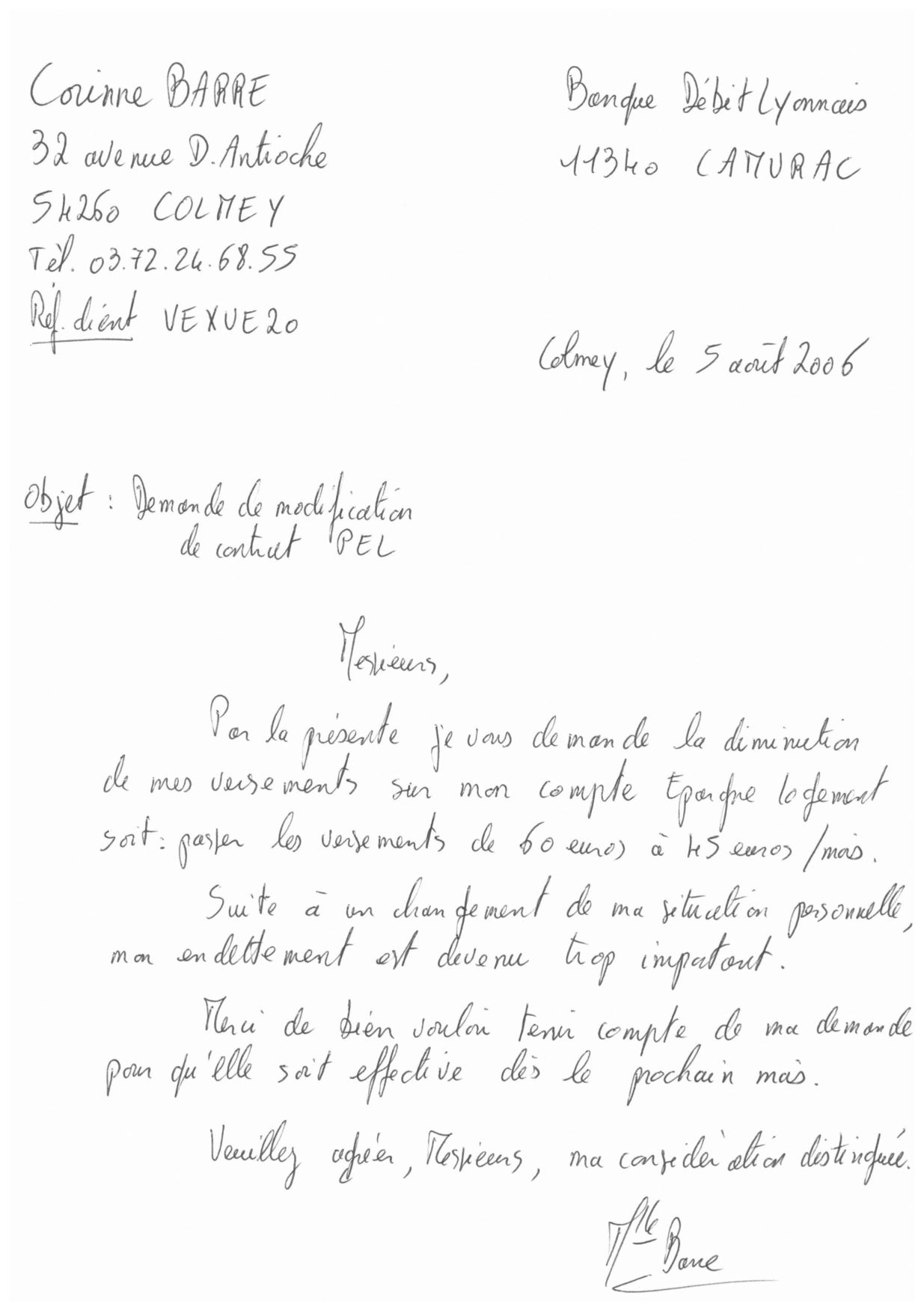}
        \caption{RIMES 2009} 
    \end{subfigure}
    \par\medskip
    \begin{subfigure}[t]{0.315\textwidth}
        \centering
        \includegraphics[width=\textwidth,frame]{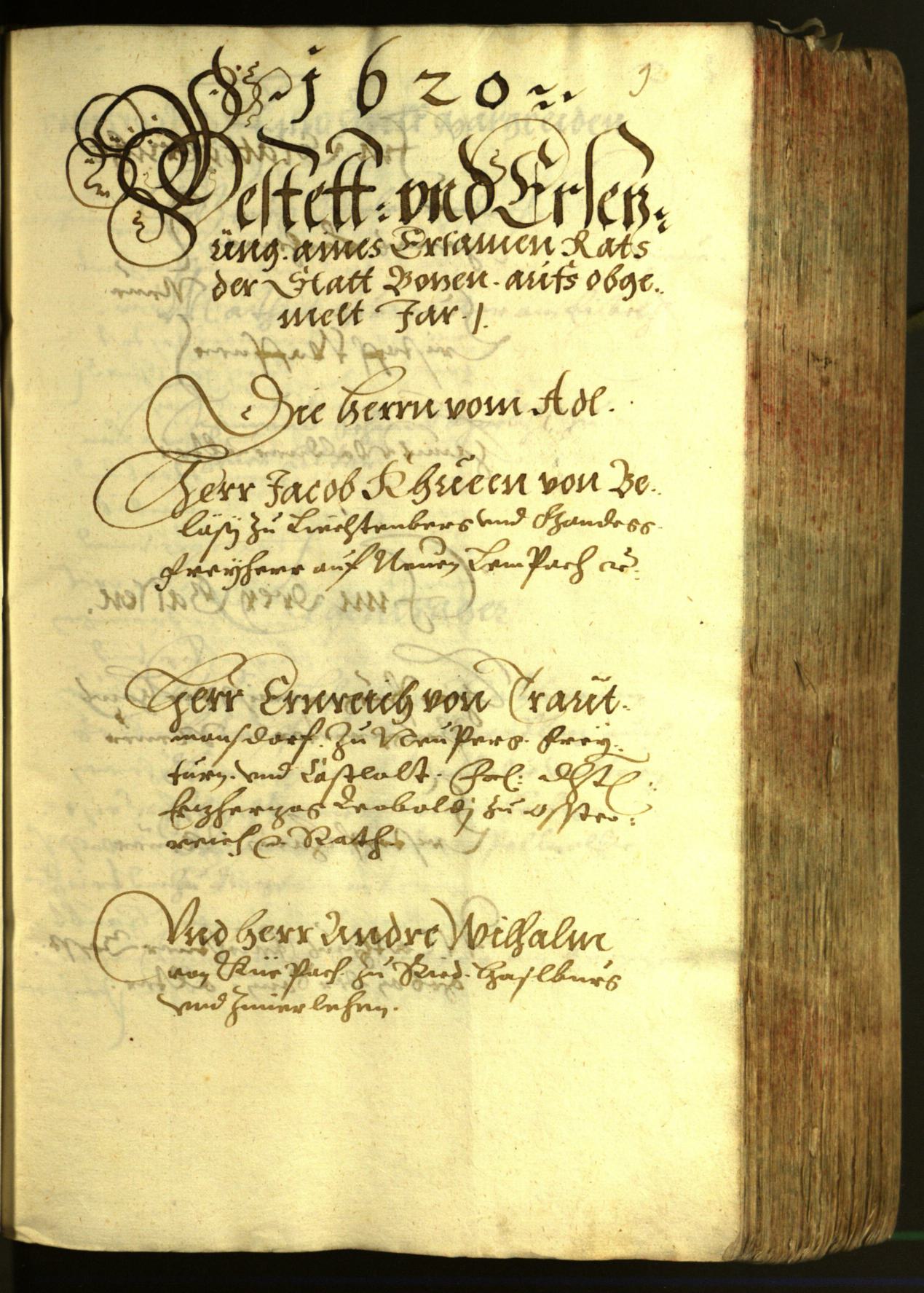}
        \caption{READ 2016 \\(single-page)} 
    \end{subfigure}
    ~
    \begin{subfigure}[t]{0.6\textwidth}
        \centering
        \includegraphics[width=\textwidth,frame]{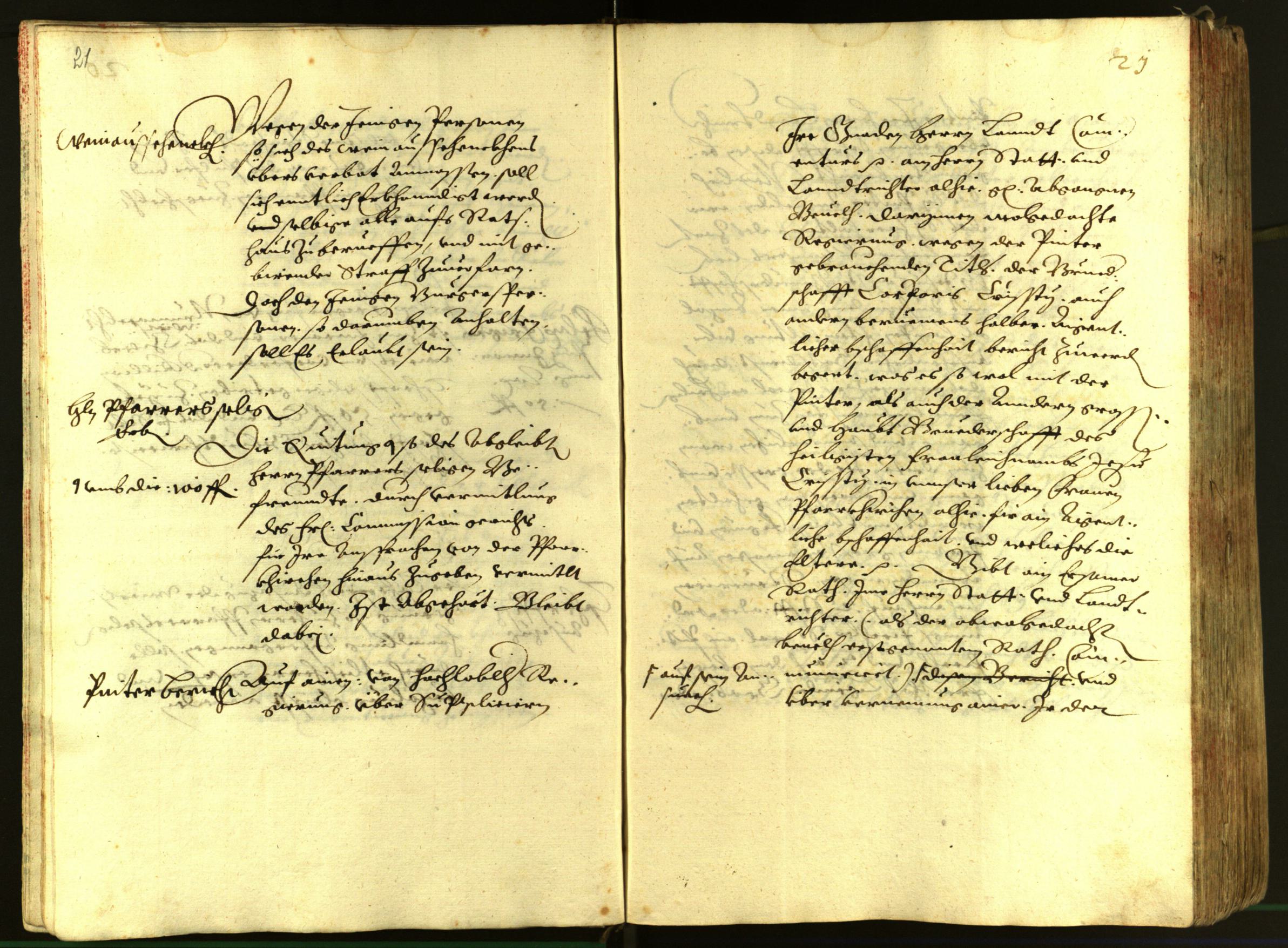}
        \caption{READ 2016 (double-page)} 
    \end{subfigure}
    \caption{Document image examples from the RIMES 2009, READ 2016 and MAURDOR datasets.}
    \label{fig:datasets}
\end{figure}

We used three document-level public datasets to evaluate the proposed approach: RIMES 2009 \cite{RIMES2009}, READ 2016 \cite{READ2016} and MAURDOR \cite{MAURDOR2014}. Document image examples from these three datasets are showed in Figure \ref{fig:datasets}.

\subsubsection{RIMES 2009}
The RIMES 2009 dataset corresponds to French grayscale handwritten page images. These pages are letters produced in the context of writing mail scenarios. Text regions are classified among one of the following 7 classes: sender coordinates, recipient coordinates, object, date \& location, opening, body and PS \& attachment. We used these classes as layout tokens. 

\subsubsection{READ 2016}
The READ 2016 dataset corresponds to Early Modern German handwritten pages from the Ratsprotokolle collection. Images are RGB encoded. We used two versions of this dataset: single-page images and double-page images. The layout classes are as follows: page, section, margin annotation and body.

\subsubsection{MAURDOR}
The MAURDOR dataset consists in a heterogeneous collection of documents. We used the same configuration as in \cite{Coquenet2022b} \textit{i.e.} we only use the English and French documents, and we focus on the C3 and C4 subsets of this dataset, which corresponds to private or professional correspondences. The documents are either handwritten, printed, or a mix of both. There is no sufficient annotation to produce the layout tokens, so we only evaluate the HTR task on this dataset.

\subsubsection{}
Table \ref{table:datasets} details the splits in training, validation and test, as well as the number of characters in the alphabet and the number of layout tokens (2 by class, for opening and closing markups) for each dataset.

\begin{table}[h]
    \caption{Splits and number of character and layout tokens for each dataset.}
    \centering
    \setlength{\tabcolsep}{6pt}
    \begin{tabular}{ l c c c c c}
    \hline
    \multirow{2}{*}{Dataset} & \multirow{2}{*}{Training} & \multirow{2}{*}{Validation} & \multirow{2}{*}{Test} & \# char & \# layout\\
    & & & & tokens & tokens\\
    \hline
    \hline
    RIMES 2009 & 1,050 & 100 & 100 & 108 & 14\\
    READ 2016 (single-page) & 350 & 50 & 50 & 89 & 10\\
    READ 2016 (double-page) & 169 & 24 & 24 & 89 & 10\\
    MAURDOR (C3) & 1,006 & 148 & 166 & 134 & \xmark\\
    MAURDOR (C4) & 721 & 111 & 114 & 127 & \xmark\\
    \hline
    \end{tabular}
    \label{table:datasets}
\end{table}

\subsection{Metrics}
In addition to the standard Character Error Rate (CER) and Word Error Rate (WER) metrics used to evaluate the text recognition performance, 
\ifanonymous
\else
we proposed two metrics in \cite{Coquenet2022b} to evaluate the specific layout recognition of the HDR task.
\fi
The Layout Ordering Error Rate (LOER) consists in considering the document layout as a graph and computing the graph edit distance between the prediction and the ground truth. The LOER aims at evaluating the layout recognition only, considering the reading order between layout entities.
Since LOER and CER/WER only evaluate the layout and text recognition independently, the $\text{mAP}_\text{CER}$ is used to evaluate the recognition of the layout with respect to the text content. It is computed as the area under the precision/recall curve, as in object detection approaches \cite{mAP_PascalVOC} for instance, but it is based on a CER threshold instead of a IoU one. The 
$\text{mAP}_\text{CER}$ does not dependent on the reading order between layout entities. That is why it is important to consider all these metrics altogether to evaluate the HDR task.

\subsection{Training details}
\ifanonymous
\else
In \cite{Coquenet2022b}, we
\fi
used some pretraining and curriculum training strategies to speed up the convergence of the DAN, and to not use any physical segmentation annotation during training.
To be fairly comparable with this work, we follow the exact same training configuration, whose major points are as follows:
\begin{itemize}
    \item The encoder is pretrained on synthetic isolated text line images using the CTC loss and a dedicated FCN line-level OCR model.
    \item The \modelname{} is trained on a mixture of real and synthetic documents. Using a curriculum strategy, the \modelname{} is trained on increasingly complex synthetic documents through the epochs. The complexity varies from two aspects: the number of lines contained in the document image, and the size of this image. The ratio between synthetic and real document also evolves during training, from 90\%/10\% to 20\%/80\%.
    \item A rule-based post-processing is used to make sure that the layout markups have the correct format (no unpaired markup, for instance).
    \item Whether it is for pretraining or training, input images are downsized to 150 dpi, normalized and data augmentation is performed 90\% of the time.
\end{itemize}
We carried out 2-day pretraining and 4-day training on a single GPU V100 (32 Go), using automatic mixed-precision. We used the Adam optimizer with an initial learning rate of $10^{-4}$. We do not use any external data, external language model nor lexicon constraints.

\subsection{Comparison with the state of the art}

To our knowledge, the only work performing HDR is the DAN \cite{Coquenet2022b}. 
Tables \ref{table:sota-read}, \ref{table:sota-rimes} and \ref{table:sota-maurdor} provide an evaluation of the \modelname{} on the READ 2016, RIMES 2009 and MAURDOR datasets, respectively, as well as a comparison with the state of the art.

\begin{table}[h!]
    \setlength{\tabcolsep}{3pt}
    \caption{Evaluation of the \modelname{} on the test set of the READ 2016 dataset and comparison with the state of the art. Metrics are expressed in percentages.}
    \centering
    \resizebox{\linewidth}{!}{
 \begin{tabular}{ l | c c c c | c c c c}
    \hline
    \multirow{2}{*}{Architecture} & \multicolumn{4}{c|}{READ 2016 (single-page)} & \multicolumn{4}{c}{READ 2016 (double-page)} \\
      & CER $\downarrow$ & WER $\downarrow$ & LOER $\downarrow$ & $\mathrm{mAP}_\mathrm{CER}$ $\uparrow$ & CER $\downarrow$ & WER $\downarrow$ & LOER $\downarrow$ & $\mathrm{mAP}_\mathrm{CER}$ $\uparrow$\\ 
    \hline
    \hline
    DAN \cite{Coquenet2022b} & \textbf{3.43} & \textbf{13.05} & 5.17 & 93.32 & \textbf{3.70} & \textbf{14.15} & 4.98 & 93.09\\
    \modelname{} & 3.95 & 14.06 & \textbf{3.82} & \textbf{94.20} & 3.88 & 14.97 & \textbf{3.08} & \textbf{94.54}\\
    \hline
    \end{tabular}
    }
    \label{table:sota-read}
\end{table}

\begin{table}[h!]
    \caption{Evaluation of the \modelname{} on the test set of the RIMES 2009 dataset and comparison with the state of the art. Metrics are expressed in percentages.}
    \centering
    \resizebox{0.6\linewidth}{!}{
    \begin{tabular}{ l | c c c c }
    \hline
    \multirow{2}{*}{Architecture} & \multicolumn{4}{c}{RIMES 2009}\\
      & CER $\downarrow$ & WER $\downarrow$ & LOER $\downarrow$ & $\mathrm{mAP}_\mathrm{CER}$ $\uparrow$ \\
    \hline
    \hline
    DAN \cite{Coquenet2022b} & \textbf{4.54} & \textbf{11.85} & \textbf{3.82} & \textbf{93.74} \\
    \modelname{} & 6.38 & 13.69 & 4.48 & 91.00\\
    \hline
    \end{tabular}
    }
    \label{table:sota-rimes}
\end{table}

\begin{table}[h!]
    \setlength{\tabcolsep}{3pt}
    \caption{Evaluation of the \modelname{} on the test set of the MAURDOR dataset and comparison with the state of the art. Metrics are expressed in percentages.}
    \centering
    \resizebox{0.75\linewidth}{!}{
    \begin{tabular}{ l | c c | c c | c c}
    \hline
    \multirow{2}{*}{Architecture} & \multicolumn{2}{c|}{C3} & \multicolumn{2}{c|}{C4 } & \multicolumn{2}{c}{C3 \& C4}\\
      & CER $\downarrow$ & WER $\downarrow$  & CER $\downarrow$ & WER $\downarrow$ & CER $\downarrow$ & WER $\downarrow$ \\
    \hline
    \hline
    DAN \cite{Coquenet2022b} & \textbf{8.62} & \textbf{18.94} & \textbf{8.02} & \textbf{14.57} & 11.59 & 27.68\\
    \modelname{} & 8.93 & 19.00 & 9.88 & 16.52 & \textbf{10.50} & \textbf{19.64}\\
    \hline
    \end{tabular}
    }
    \label{table:sota-maurdor}
\end{table}

The \modelname{} reaches competitive results compared to the DAN on the three datasets. For the READ 2016 dataset, it even reaches state-of-the-art results in terms of LOER and $\mathrm{mAP}_\mathrm{CER}$ for both single-page and double-page versions, involving a better recognition of the layout. Results are not as good for the RIMES 2009 dataset, which includes more variability in terms of layout. We assume that this higher variation makes the first pass of the \modelname{} more difficult. This is confirmed when measuring the CER for the first pass only: it is of 4.72\% and 5.34\% for READ 2016 at single-page and double-page levels, and of 9.10\% for RIMES 2009. Concerning the MAURDOR dataset, the \modelname{} reaches competitive results on the C3 and C4 categories, taken separately, and it reaches new state-of-the art results when mixing both categories with 10.50\% of CER, compared to 11.59\% for the standard DAN.

\subsubsection*{Discussion}
It has to be noted that it is more difficult to compare the text recognition performance at document level than at line level. Indeed, the reading order is far more complex for documents, to go from one paragraph to another, and to one line to the next, than for isolated lines. This way, even perfectly recognized, the CER can be severely impacted if the paragraphs are recognized in the wrong order. On the contrary, the $\mathrm{mAP}_\mathrm{CER}$ is invariant to the order of the layout entities, but it is dependent to the well recognition of the layout. 

Another point to emphasize is about the severity of the errors made. There are two types of errors to be distinguished. The first corresponds to standard character addition, removal, or substitution cases. During the first pass of the \modelname{}, this kind of error may have a great impact because a whole text line may be duplicated or discarded. However, during the second pass, we assume that the impact of such errors is rather equivalent for both DAN and \modelname{}. The second kind of errors is related to the end-of-transcription token prediction. Indeed, although rare, the model may not predict the end of the transcription and loop on the same text region again and again until reaching an arbitrary chosen iteration limit. For this later issue, the standard DAN is more impacted than the \modelname{}. Indeed, the DAN only have one iteration limit, which corresponds to the global number of tokens to predict for the whole document. For the \modelname{}, we used two iteration limits: one for the number of lines, and one for the number of characters per line. Given that the range of values for a line length is smaller than for the whole document, the impact is less important for the \modelname{}. 

\subsubsection*{Prediction time}
Table \ref{table:times} shows a comparison of the \modelname{} with the DAN in terms of prediction times for the three datasets: RIMES 2009, READ 2016 and MAURDOR. To be fairly comparable, these times account for the whole prediction process, including the time dedicated to the encoder part and to formatting instructions. Additional details are given for each dataset such as the image sizes, the number of characters, lines, and layout tokens per image, and the number of characters per line. The values are given as average for the test set of each dataset. As one can note, the \modelname{} is significantly faster than the DAN for all the datasets, speeding up the prediction process by at least 4. 

\begin{table}[h!]
    \setlength{\tabcolsep}{3pt}
    \caption{Prediction time comparison between the DAN and the \modelname{}. Times (in seconds) are averaged on the test set for a single document image, using a single GPU V100.}
    \centering
    \resizebox{\linewidth}{!}{
    \begin{tabular}{ l | c | c | c | c | c | c }
    \hline
   & RIMES 2009 & \multicolumn{2}{c|}{READ 2016} & \multicolumn{3}{c}{MAURDOR}\\
      & & single-page & double-page & C3 & C4 & C3 \& C4 \\
    \hline
    \multicolumn{7}{l}{Dataset details (averaged for a document on the test set)}\\
    \hline
    width (px) & 1,235 & 1,190 & 2,380 & 1,336 & 1,240 & 1,297 \\
    height (px) & 1,751 & 1,755 & 1,755 & 1,658 & 1,754 & 1,697 \\
    \# chars & 578 & 528 & 1,062 & 481 & 706 & 575\\
    \# lines & 18 & 23 & 47 & 16 & 22 & 18\\
    \# chars / line & 31 & 22 & 22 & 30 & 31 & 30\\
    \# layout tokens & 11 & 15 & 30 & 0 & 0 & 0\\
    \hline
    \multicolumn{7}{l}{Prediction times (in seconds)}\\
    \hline
    DAN \cite{Coquenet2022b} & 5.6 & 4.6 & 8.5 & 5.8 & 7.7 & 6.6 \\
    \modelname{} & \textbf{1.4} & \textbf{0.9} & \textbf{1.9} & \textbf{1.0} & \textbf{1.6} & \textbf{1.3}\\
    \hline
    Speed factor & x4 & x5.1 & x4.5 & x5.8 & x4.8 & x5.1 \\
    \hline
    \end{tabular}
    }
    \label{table:times}
\end{table}

We showed that the \modelname{} reaches competitive results on three document-level datasets while being at least 4 times faster than the standard DAN at prediction time. We now evaluate the performance on heterogeneous documents, by mixing both RIMES 2009 and READ 2016 datasets.

\subsection{Evaluation on heterogeneous documents}

In this experiment, we mixed both RIMES 2009 and READ 2016 datasets at single page level, for both training and evaluation. Examples from both datasets are balanced at training time, \textit{i.e.}, the models have been trained on the same number of documents for both datasets. 
These are the first results for such an experiment; we also train the standard DAN for comparison purposes. Results are shown in Table \ref{table:cross-rimes-read}. As one can note, results are rather similar when training on datasets separately or altogether, except for the DAN on the RIMES dataset whose CER increases from 4.54\% up to 7.96\%.

\begin{table}[h!]
    \setlength{\tabcolsep}{3pt}
    \caption{Evaluation of the \modelname{} on heterogeneous data (mixing READ 2016 and RIMES 2009 for both training and evaluation) and comparison with the state of the art.}
    \centering
    \resizebox{\linewidth}{!}{
 \begin{tabular}{ l | c c c c | c c c c}
    \hline
    \multirow{2}{*}{Architecture} & \multicolumn{4}{c|}{RIMES 2009 (page)} & \multicolumn{4}{c}{READ 2016 (single-page)} \\
      & CER $\downarrow$ & WER $\downarrow$ & LOER $\downarrow$ & $\mathrm{mAP}_\mathrm{CER}$ $\uparrow$ & CER $\downarrow$ & WER $\downarrow$ & LOER $\downarrow$ & $\mathrm{mAP}_\mathrm{CER}$ $\uparrow$\\ 
    \hline
    \hline
    DAN \cite{Coquenet2022b} & 7.96 & 15.76 & 8.72 & \textbf{91.59} & \textbf{3.50} & \textbf{13.36} & \textbf{3.86} & \textbf{94.23}\\
    \modelname{} & \textbf{6.73} & \textbf{15.22} & \textbf{5.56} & 90.10 & 3.81 & 14.30 & 4.32 & 93.57\\
    \hline
    \end{tabular}
    }
    \label{table:cross-rimes-read}
\end{table}

\subsection{Ablation study}
\label{section:ablation}

In Table \ref{table:fdan-ablation}, we propose an ablation study of the proposed approach on the RIMES 2009 and READ 2016 datasets. The first line corresponds to the \modelname{} baseline. In experiment (1), the document positional encoding is replaced by standard 1d positional encoding, \textit{i.e.}, a unique index is associated to each token. The model does not succeed to recognize the text, showing the necessity of injecting line positional information to parallelize the recognition.
The model can only access to tokens of the text line to recognize in (2), also preventing the text recognition. Indeed, it is nearly impossible to predict the next character with only a one-character query (beginning of the second pass) since characters are not unique in a document. 
For both experiments, one can note that the LOER is nearly not impacted, this is because the layout recognition takes place in the first pass, before the parallelization.

\begin{table}[t!]
    \setlength{\tabcolsep}{3pt}
    \caption{Ablation study of the \modelname{} and DAN. Results (in percentages) are given for the test set of the RIMES 2009 and READ 2016 datasets.}
    \centering
    \resizebox{\linewidth}{!}{
 \begin{tabular}{ l | c c c | c c c | c c c}
    \hline
    \multirow{2}{*}{Architecture} & \multicolumn{3}{c|}{RIMES 2009 (page)} & \multicolumn{3}{c|}{READ 2016 (single-page)} & \multicolumn{3}{c}{READ 2016 (double-page)}\\
      & CER & LOER & $\mathrm{mAP}_\mathrm{CER}$ & CER & LOER & $\mathrm{mAP}_\mathrm{CER}$ & CER & LOER & $\mathrm{mAP}_\mathrm{CER}$\\ 
    \hline
    \modelname{} & \textbf{6.38} & 4.48 & 91.00 &  3.95 & \textbf{3.82} & \textbf{94.20} & \textbf{3.88} & \textbf{3.08} & \textbf{94.54}\\
    (1) No line encoding & 79.39 & 6.21 & 0.00 & 75.08 & 11.81 & 0.29 & 75.01 & 10.79 & 5.44\\
    (2) Single-line context & 94.73 & \textbf{4.30} & 0.00 & 91.23 & 4.61 & 0.00 & 91.22 & 4.03 & 0.00\\
    (3) First-pass context & 8.27 & 4.90 & 90.73 & 6.68 & 4.50 & 88.37 & 6.87 & 5.22 & 87.93\\
    (4) Sum PE & 6.88 & 4.90 & \textbf{91.06} & \textbf{3.82} & 4.27 & 94.08 & 4.55 & 4.39 & 92.76\\
    \hline
    \end{tabular}
    }
    \label{table:fdan-ablation}
\end{table}

In experiment (3), in addition to the tokens of the text line to recognize, the first character of all the text lines, as well as the layout markup tokens, are available. This leads to an increase of the CER of at least 1.89 points for RIMES 2009, and up to 2.99 points for READ 2016 at double-page level, compared to the baseline. This shows the efficiency of the text line detection performed in the first pass, since the text recognition is parallelized, but it also demonstrates that gathering the context from past and future lines helps to improve the performance.
In experiment (4), the positional encoding of the line and of the index in the line are summed instead of being concatenated. As one can note, results are slightly in favor of the concatenation. 

\section{Conclusion}
\label{conclusion}
In this paper, we proposed the \modelname{}, a novel approach for end-to-end Handwritten Document Recognition.  We evaluate this approach with the current state-of-the-art architecture and showed that this approach reaches competitive results on three document-level datasets while being at least 4 times faster. This way, we preserved the advantages of using a single end-to-end approach, while greatly mitigating the major drawback of prediction time. 
In this work, we focused on line-level multi-target queries to show the gain in prediction time. However, it would also be possible to perform this parallelization at paragraph level in order to have a more important language modeling of the past: this would represent an in-between in terms of prediction time.

\ifanonymous
\else
\subsubsection*{Acknowledgments}
This work was granted access to the HPC resources of IDRIS under the allocation 2020-AD011012155.
\fi

\bibliographystyle{splncs04}
\bibliography{references.bib}

\end{document}